%% file: main.tex
\documentclass[10pt,twocolumn,letterpaper]{article}

\usepackage[pagenumbers]{cvpr} 


\definecolor{cvprblue}{rgb}{0.21,0.49,0.74}
\usepackage[pagebackref,breaklinks,colorlinks,allcolors=cvprblue]{hyperref}

\usepackage[utf8]{inputenc} 
\usepackage[T1]{fontenc}    
\usepackage[accsupp]{axessibility}  

\usepackage{algorithm}
\usepackage{amsmath}
\usepackage{booktabs}
\usepackage{graphicx}
\usepackage{here}
\usepackage{multirow}
\usepackage[all]{nowidow}
\usepackage{uoftcolors}
\usepackage{abbreviations}
\usepackage{verbatim}
\usepackage{algpseudocode}
\usepackage{nicefrac}       
\hypersetup{
  colorlinks=true,
  allcolors=uoftoceanblue
}

\def\paperID{12154} 
\def\confName{CVPR}
\def\confYear{2025}

\title{EventSplat: 3D Gaussian Splatting from Moving Event Cameras for Real-time Rendering}

\author{Toshiya Yura$^\text{1,2}$\\
{\tt\small toshiya.yura@sony.com}
\and
Ashkan Mirzaei$^\text{2}$\\
{\tt\small ashkan@cs.toronto.edu}
\and
Igor Gilitschenski$^\text{2}$\\
{\tt\small gilitschenski@cs.toronto.edu}\\
$^\text{1}$Sony Semiconductor Solutions Corporation, $^\text{2}$University of Toronto\\
}

\begin{document}
\maketitle
\input{sec/0_abstract}    
\input{sec/1_intro}
\input{sec/2_related_work}
\input{sec/3_background}
\input{sec/4_method}
\input{sec/5_experiments}
\input{sec/6_conclusion_and_futurework}

\section*{Acknowledgments}
The work is in part supported by Sony. We would also like to thank Ziyi Wu, Yash Kant, and Umangi Jain, for valuable discussions and support.

{
    \small
    \bibliographystyle{ieeenat_fullname}
    \bibliography{main}
}

\input{sec/appendix}

\end{document}

%% file: sec/0_abstract.tex
\begin{abstract}
We introduce a method for using event camera data in novel view synthesis via Gaussian Splatting.
Event cameras offer exceptional temporal resolution and a high dynamic range. 
Leveraging these capabilities allows us to effectively address the novel view synthesis challenge in the presence of fast camera motion.
For initialization of the optimization process, our approach uses prior knowledge encoded in an event-to-video model. We also use spline interpolation for obtaining high quality poses along the event camera trajectory. 
This enhances the reconstruction quality from fast-moving cameras while overcoming the computational limitations traditionally associated with event-based Neural Radiance Field (NeRF) methods. Our experimental evaluation demonstrates that our results achieve higher visual fidelity and better performance than existing event-based NeRF approaches while being an order of magnitude faster to render.
\end{abstract}

\begin{figure*}[t]
    \centering
    \includegraphics[width=0.8\linewidth]{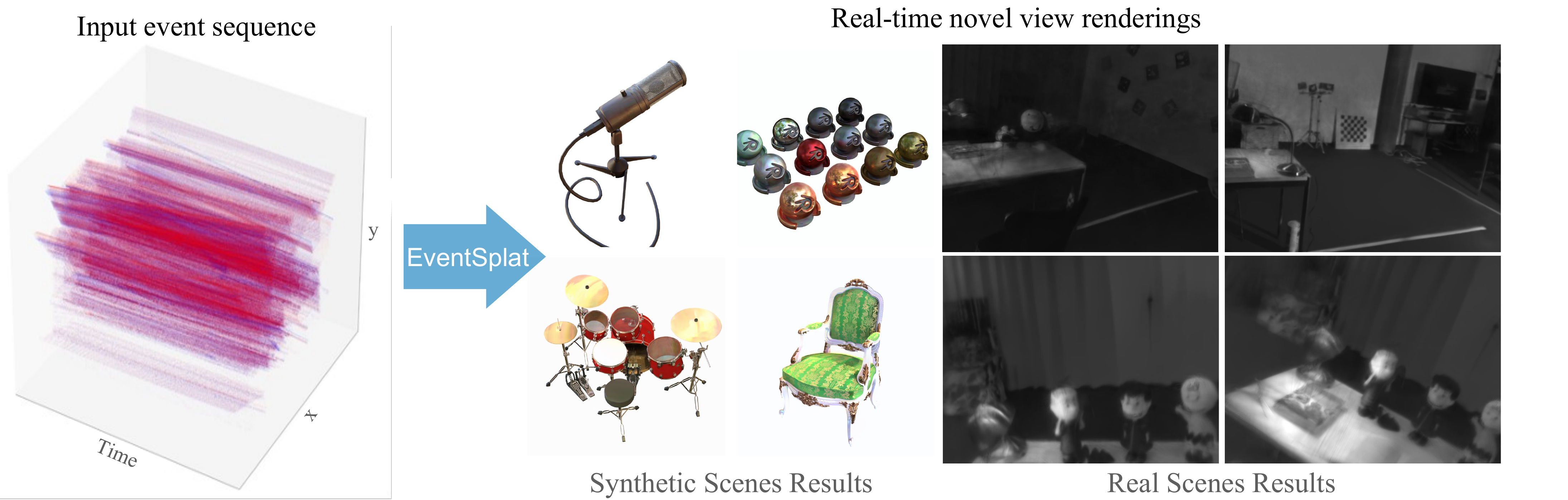}
    \caption{EventSplat derives 3D representations of scenes in the form of 3D Gaussians from event data, enabling fast real-time renderings of scenes captured with event cameras via rasterization of 3D Gaussians. The use of event-based input is particularly advantageous when traditional RGB cameras fail due to various reasons including poor lighting conditions or motion blur from fast-moving cameras.}
    \label{fig:teaser}
\end{figure*}

%% file: sec/1_intro.tex
\section{Introduction}
\label{sec:intro}
Event cameras are a vision modality that draws strong inspiration from biological vision and neuromorphic design. In these sensors, each pixel operates independently by triggering events upon a log-intensity change above a pre-defined threshold~\cite{4444573,9138762}. This results in an asynchronous data stream as opposed to capturing absolute intensity values at fixed intervals in traditional cameras. Event cameras offer low latency, high temporal resolution, and a high dynamic range while also ensuring minimal energy consumption~\cite{10341804,10.1007/978-3-030-58598-3_41}. Consequently, they are particularly suitable for dynamic and challenging lighting conditions, as well as environments with rapid camera motions where classical cameras may exhibit strong motion blur~\cite{10.1007/978-3-030-58598-3_41,Yu_2021_ICCV,Tulyakov_2022_CVPR,Tulyakov_2021_CVPR}. These advantages have motivated the adoption of event cameras in a wide range of applications covering most of the traditional computer vision tasks~\cite{Gallego_2018_CVPR,doi:10.1126/scirobotics.aaz9712,kim2016real,Pan_2019_CVPR,osswald2017spiking}.

Recently, novel view synthesis has seen significant advances. This progress has been particularly driven by the emergence and development of Neural Radiance Fields (NeRF)~\cite{10.1145/3503250,mipnerf,mipnerf.360,yariv2023bakedsdf,barron2023zipnerf,Lassner_2021_CVPR}.
Recent works have led to NeRF improvements in terms of rendering speed, and adoption of NeRFs in 3D editing and other downstream applications~\cite{Ma_2022_CVPR,Mildenhall_2022_CVPR,zhu2021autorecon,Mirzaei_2023_CVPR,muller2022instant,mobile.nerf}.
3D Gaussian Splatting (3DGS)~\cite{kerbl20233d,wu20234d} achieves real-time rendering by replacing the underlying representation with 3D Gaussians and using fast differentiable rasterization. The substantial speed improvement of 3DGS is achieved without compromising the visual quality.

The success of NeRFs in view-synthesis has led to their adoption in applications with event-based cameras~\cite{rudnev2023eventnerf,klenk2023nerf,hwang2023ev,low2023robust}.
Conventional NeRF approaches struggle with challenges, such as 3D reconstruction in environments with rapidly changing lighting conditions or high-speed objects due to RGB camera limitations.
The integration of NeRFs with event-based cameras leverages their unique advantages and incorporates them into the NeRF framework to address the challenges that conventional NeRF approaches face.

However, one crucial area of research focuses on expediting the rendering processes to fully unlock the potential of event-based NeRFs. Leveraging the advantages of 3D Gaussian Splatting with events requires adapting the rasterization process to allow supervision with events or accumulated events rather than RGB images.


In summary, this paper contributes a method for 3D Gaussian Splatting from event data by leveraging the following technical ideas:
\begin{enumerate}[label=\arabic*)]
  \item An integration of event accumulation with the optimization process of 3DGS.
  \item An event-to-video guided Structure from Motion (SfM) approach for initializing the 3DGS optimization process.
  \item Use of cubic spline trajectory interpolation to assign camera poses to events at high rates.
\end{enumerate}
Our evaluations show that these design choices achieve state-of-the-art event-based novel view synthesis performance while benefiting from fast rendering speeds. Thus, our method represents an advancement in addressing the speed constraints associated with event-based NeRFs.
It opens up new possibilities for real-time novel view synthesis for environments where traditional imaging techniques struggle.

%% file: sec/2_related_work.tex
\section{Related Work}
\label{sec:relatedwork}
\subsection{Event Cameras for Vision Applications}
Motivated by the low energy consumption, high dynamic range, and high asynchronous temporal resolution of event cameras, event-based vision has recently garnered significant attention~\cite{9138762}. These advantageous properties have led to promising advancements in downstream tasks such as object tracking, detection, and recognition~\cite{EventTracking1,EventTracking3,EKLT,1MpxDet,RVT,ASTMNet,EST,DiST-N-IN,HATS-N-Cars,wu2023leod}, as well as optical flow estimation~\cite{EV-FlowNet,E-RAFT,EventOptFlow1}. However, being a novel data format, event data necessitates specialized architectures and algorithms to be effectively processed and utilized in various applications. In this work, we present a method to exploit event data for real-time novel view synthesis, a scenario that has not been addressed previously.

\subsection{Volumetric Rendering for View Synthesis}
Neural Radiance Fields (NeRFs)\cite{10.1145/3503250} have been the state-of-the-art in view synthesis, enabled by volumetric rendering and neural scene representations. Significant research has been conducted to improve NeRFs in terms of training and rendering speed~\cite{plenoxels,chen2022tensorf,muller2022instant,plenoctrees,mobile.nerf,Hedman_2021_ICCV,merf,kurz2022adanerf} and visual quality~\cite{ds.nerf,mipnerf,mipnerf.360,bacon,refnerf}. However, due to the expensive volumetric sampling process, NeRF-based methods have generally been unable to render high-quality views in real time. Recently, 3D Gaussian Splatting (3DGS)~\cite{chen2024survey,kerbl20233d,964490,chen2023textto3d,poole2022dreamfusion} proposed a rasterization-based approach for differentiably rendering scenes represented as sets of 3D Gaussian ellipsoids. This method achieves real-time renderings while preserving the quality of state-of-the-art NeRF-based methods.
Since the introduction of 3DGS, real-time capabilities of 3DGS have resulted in rapid adoption of Gaussians in several downstream tasks including dynamic scene reconstruction~\cite{yang2023gs4d}, 3D asset generation~\cite{yi2023gaussiandreamer}, SLAM~\cite{gsslam}, and 4D object generation~\cite{ling2023align}.
For view synthesis from event data, we leverage 3DGS and adapt it to be trainable using event sequences.

\subsection{Novel View Synthesis via Event Data}
Motivated by the unique benefits of event-based cameras, such as high dynamic range, low latency, and high temporal resolution, there has been significant research interest in adopting event data in applications like 3D reconstruction and novel view synthesis in cases where traditional RGB cameras underperform~\cite{Zhou_2018_ECCV,rebecq2016evo,zhou2021event,9138762}.
Recent methods explore using NeRFs with event-vision supervision~\cite{rudnev2023eventnerf,klenk2023nerf,hwang2023ev,low2023robust} to reconstruct scenes and render them from novel views. While capable of generating high-quality renderings, NeRF-based methods are generally slow to render due to expensive sampling requirements~\cite{chen2022tensorf,garbin2021fastnerf,reiser2021kilonerf,takikawa2021neural}. On the other hand, 3DGS requires RGB data for supervision, and adapting 3DGS for scenarios where the supervision signal comes from an event-based camera has been challenging~\cite{wang2024evggs,yu2024evagaussians,xiong2024event3dgs,deguchi2024e2gs,wu2024ev,weng2024eadeblur,zhang2024elite}. In contrast to previous event-based view synthesis methods, we enable the use of event data to supervise 3DGS, allowing for real-time renderings and state-of-the-art quality using only event data, supported by event-based initial point clouds and effective camera trajectory interpolation.

%% file: sec/3_background.tex
\section{Background}
\label{sec:background}
\subsection{Event Camera Data}
While traditional cameras output an intensity image $I$, event cameras model changes in the log intensity $L=\log(I)$ of the current camera view. Similar to traditional cameras, they have a pixel array. However, in contrast to traditional cameras, each pixel acts as an independent sensor and triggers an event whenever the difference between the current log-intensity \(l_{x,y}(t)\) and the log-intensity at the time of the most recent event \(l_{x,y}(t_{\text{recent}})\) surpasses a given threshold $\Delta>0$, i.e. when
\begin{equation}
\label{eq:event_trigger}
  l_{x,y}(t) - l_{x,y}(t_{\text{recent}}) > \Delta.
\end{equation}
This threshold is also known as \emph{contrast sensitivity}~\cite{9138762}. It can be different for positive and negative polarity changes.

Formally, each event is a tuple $e:=(t, x, y, p)$ where $t\in\R$ is a timestamp at which it was triggered, \((x,y)\in\Z^2\) are its pixel coordinates, and \(p \in \{-1,\,1\} \) is the event's polarity indicating whether the log-intensity increased or decreased. Similar to traditional color cameras, a Bayer RGB pattern is used to obtain color information in color event cameras. In such sensors, each individual event provides information only for one color channel.

%
%

\subsection{3D Gaussian Splatting}
To enable novel view synthesis, 3DGS learns a scene representation that consists of a set of ellipsoids with view-dependent color information encoded as spherical harmonics~\cite{chen2024survey,kerbl20233d,964490,chen2023textto3d,poole2022dreamfusion}.
The essence of the method revolves around transforming optimized 3D Gaussians into 2D imagery based on predetermined camera positions
Subsequently the approach calculates pixel values accordingly by arranging and rasterization~\cite{964490,kerbl20233d,chen2024survey,chen2023textto3d,poole2022dreamfusion}. The transformation process involves mapping 3D Gaussian distributions, shaped like ellipsoids, onto a two-dimensional plane as ellipses for the purpose of rendering. This is achieved by applying a specific viewing transformation, denoted by $W$, along with a 3D covariance matrix $\Sigma$. The outcome is the derivation of a 2D covariance matrix $\Sigma'$ as follows
\begin{equation}
\label{eq:2d_covariance}
  \Sigma' = JW\Sigma W^{\top}J^{\top},
\end{equation}
where the Jacobian $J$ is used to linearly approximate projective transformations.

The pixel rendering procedure begins by pinpointing a pixel's location, represented as $p$, within the image plane and then assess its closeness to the intersecting Gaussians. This step involves gauging the Gaussian's depth via applying a viewing transformation, denoted as $W$~\cite{chen2024survey}. This process forms an sorted sequence of Gaussians, labeled~$N_s$. Subsequently, a technique known as alpha compositing is employed to determine the pixel's color
\begin{equation}
\label{eq:color}
  I = \sum_{i \in N_s} \left( c_i \alpha'_i \prod_{j=1}^{i-1} (1 - \alpha'_j) \right).
\end{equation}
Here, $c_i$ and $\alpha_i'$ indicates the learned color and the opacity, respectively.

%% file: sec/4_method.tex
\section{Method}
\label{sec:method}
The goal of our work is to obtain a Gaussian Splatting-based 3D representation of a static scene using a data stream from a moving event camera. Formally, we are given a sequence of $N$ events \(e_k=(t_k, x_k, y_k, p_k)\) with corresponding camera poses $P_k\in \text{SE}(3)$ denoted as \(\mcE:=\{(e_k, P_k)\}_{k=1}^N\). In practice, such dense pose information may not be available. However, it can be computed from sparse poses via interpolation. The interpolation to estimate event camera poses is explained in~\cref{sec:scenes}. From the posed event sequence, we aim to obtain a 3DGS-based scene representation $\mcG$ that can be used for novel view synthesis of traditional rgb and grayscale images. 

The key idea of our optimization approach is to model changes in logarithmic image intensity. That is, for a trajectory sequence over a given time interval $[a, b]$ and corresponding image intensity $I(t)\in\R^{H\times W \times 3}$, our work models 
\begin{equation}
\label{eq:logInteistyChanges}
  E(a,b):=\int_{a}^{b} \log\left( I'(t) \right) \mathrm{d} t .
\end{equation}
For loss computation, we use both, event camera data and 3DGS to approximate this quantity. In the case of event cameras, this approximation can be obtained directly by creating an image of log intensity changes. Approximating~\cref{eq:logInteistyChanges} with 3DGS, however, requires additional modifications to the training process.

\paragraph{Method Overview}

\begin{figure*}[tb]
  \centering
  \includegraphics[width=15.0cm]{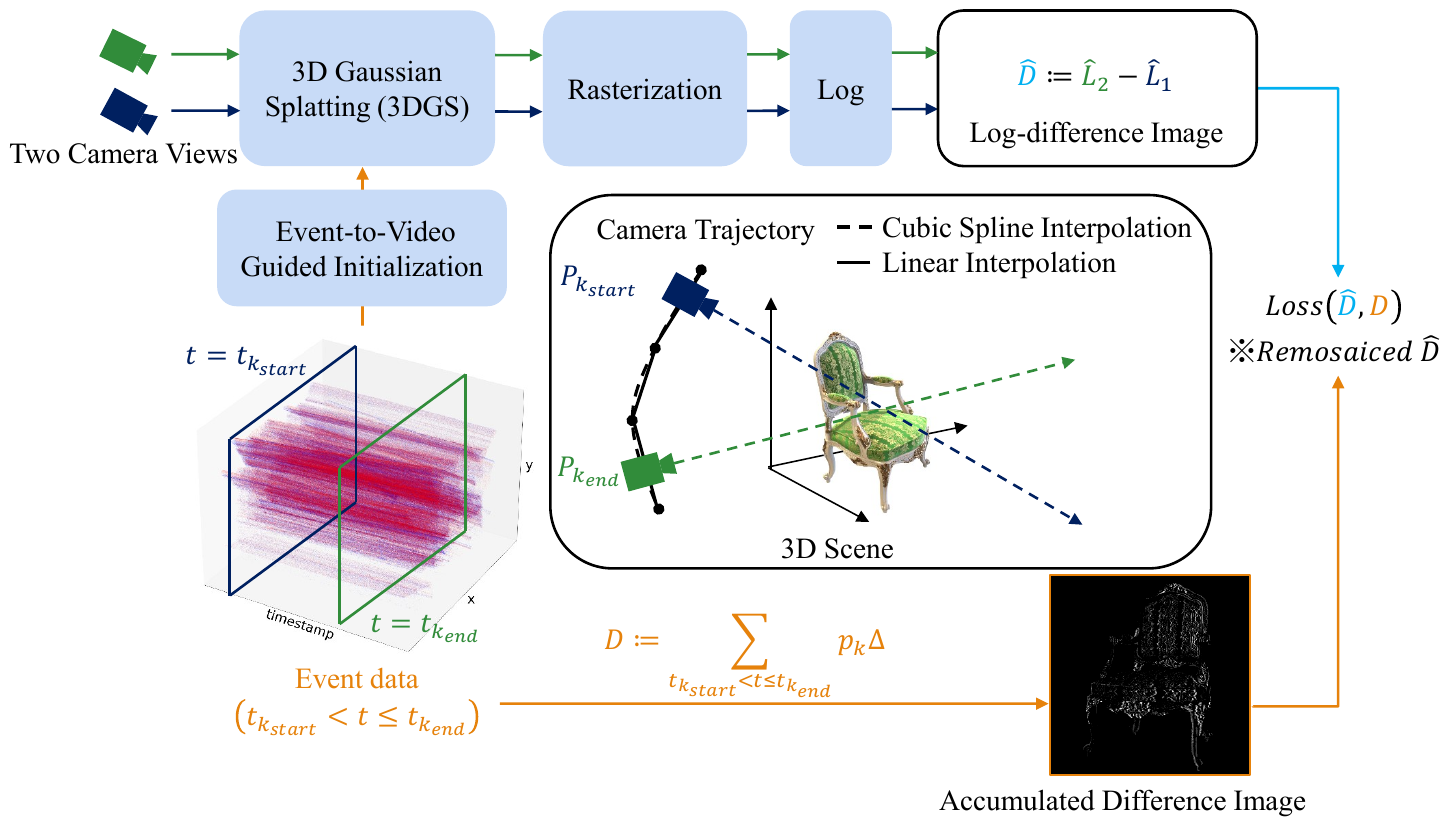}
  \caption{Overview of our 3D Gaussian Splatting training with moving event camera data. Event data streams from~$t_{k_{start}}$ to~$t_{k_{end}}$ are accumulated into~$D$, and distill the point priors from an event-to-video model. The log-difference image~$\hat{D}$ is obtained from 3D Gaussian Splatting and Rasterization at two camera view points. It is computed as in~\cref{eq:log_diff} and remosaicing is performed in case of a color event camera. The respective poses are estimated by cubic spline interpolation.}
  \label{fig:overview}
\end{figure*}

In its original formulation, 3D Gaussian Splatting uses a sequence of posed images as its training data. To make Gaussian Splatting applicable to event data, we introduce several modifications to the original method: First, we approximate~\cref{eq:logInteistyChanges} by accumulating events into an image representation $D\in\R^{H\times W}$ using a random sub-trajectory of the given event stream \(\mcE\) (\cref{sec:event_accumulation}). This representation contains the sum of quantized log-intensity changes over the sub-trajectory. Second, we compute the corresponding log-intensity changes by rasterizing two views from the Gaussian scene representation (\cref{sec:viewgeneration}). For color cameras, this additionally requires a remosaicing step to account for the fact that the accumulated image representation of events only contains single color information for each pixel. To improve the 3DGS optimization process, we leverage an event-to-video guided structure from motion (\cref{sec:e2vidinit}). We also leverage cubic spline interpolation to more accurately estimate camera sub-trajectories as the rate at which the camera is tracked may be too coarse (\cref{sec:interpolation}). The full scene representation learning process is illustrated in~\cref{fig:overview}.

\subsection{Event Accumulation}
\label{sec:event_accumulation}
To obtain an image of accumulated events, we first randomly sample start and endpoints \(k_{\text{start}}, k_{\text{end}}\in \{1\ldots N\} \) from our trajectory sequence. The lengths of those sequences are chosen randomly with a maximum length ranging from 1\% to 10\% of the total number of events. This allows for considering scene information at different scales in order to have both global and local geometry information trained. Then, we obtain the entries of the accumulated log-intensity change image $D$ at pixel location $x,y$ as
\begin{equation}
D_{x,y} := \sum_{\substack{k\in \{k_{\text{start}},\ldots, k_{\text{end}}\}\\ x_k=x,\,y_k=y} } 
p_k \Delta\ .
\end{equation}
The resulting image serves as an approximation to~\cref{eq:logInteistyChanges}, i.e. $D\approx E(t_{\text{start}}, t_{\text{end}})$ This method can easily be adapted to the case where the event-threshold $\Delta$ is asymmetric, i.e., when it is different for positive and negative polarity events.

\subsection{Training View Generation and Remosaicing}
\label{sec:viewgeneration}
To generate the log-difference image $\hat{D}\in\R^{H\times W}$, we compute
\begin{equation}
\label{eq:log_diff}
  \hat{D} := \hat{L}_2-\hat{L}_1 
\end{equation}
which requires the remosaiced log images \(L_1, L_2\in\R^{H\times W}\) at the poses $P_{k_{\text{start}}}$ and $P_{k_{\text{end}}}$ respectively.
These poses are the same that were used for generating our target \(D\). To obtain $L_1$ and $L_2$ we first use the rasterizer to generate corresponding views $I_1, I_2\in\R^{H\times W \times 3}$ from the current Gaussian scene representation. Next, we perform a standard remosaicing operation because events are triggered per pixel asynchronously, not allowing us to use conventional demosaicing methods~\cite{malvar2004high, li2005demosaicing, longere2002perceptual, li2008image, kimmel1999demosaicing, cao2009accurate}
\begin{equation}
\label{eq:remosaicing}
  \text{Remosaicing}: \R^{H\times W\times 3} \to \R^{H \times W}
\end{equation}
which reintroduces the Bayer RGB pattern. It consists of two steps: First, a set of color-channel specific matrices of size $2\times 2$ are Hadamard multiplied with each $2\times 2$ pixel block in the image. The channel specific matrices $R,G,B\in\R^{2\times 2}$ are given by 
\begin{equation}
  R=\begin{bmatrix}
  1 & 0\\
  0 & 0
\end{bmatrix},\quad
  G=\begin{bmatrix}
  0 & 1\\
  1 & 0
\end{bmatrix},\quad
  B=\begin{bmatrix}
  0 & 0\\
  0 & 1
\end{bmatrix}.
\end{equation}
The thus modified channels have non-zero entries at non-overlapping pixel locations. From this, the single-channel remosaiced image is obtained by addition across the channels. Finally, entry-wise logarithm computation yields the desired images \(\hat{L}_1\), \(\hat{L}_2\) and thus $\hat{D}$.

\subsection{Event-to-Video Guided Initialization}
\label{sec:e2vidinit}
Event cameras capture only the pixels that indicate changes in brightness. This results in sparse information that is often insufficient for effective supervision. Thus, reconstructing 3D information, including background details, solely from event data is exceedingly challenging. To simplify the optimization process we propose leveraging the prior knowledge encoded in an event-to-video model. We use this pretrained model to generate images from event streams that can subsequently be processed by Structure from Motion (SfM) techniques. Although the generated images often contain significant noise and do not accurately reproduce the underlying RGB image values, they retain substantial texture and background information. This information is sufficient to provide initial positions for the Gaussians, improve the 3DGS optimization process and lead to faster convergence compared to basic initialization with random initial points.

\subsection{Event Camera Trajectory Interpolation}
\label{sec:interpolation}
The camera poses, $P_{k_{\text{start}}}$ and $P_{k_{\text{end}}}$, used to obtain the log images \(\hat{L}_1\) and \(\hat{L}_2\) in~\cref{eq:log_diff}, are in most cases not given a priori. This is because event camera poses are typically obtained from a tracking or motion estimation system that operates at a fixed rate, which may not be aligned with the rate of event accumulation.
Thus, to obtain an estimate of the complete trajectory of the event camera, we adopt an efficient trajectory interpolation method for camera poses. Our work utilizes Cubic Spline Interpolation and Spherical Cubic Spline Interpolation.
These generate a smooth path by interpolating between discrete command points using third-degree polynomials. This interpolation technique closely approximates real-world camera motion by maintaining non-linear continuity in both velocity and acceleration of camera movement.

\subsection{Loss Function}
3D Gaussian Splatting is updated using a loss that combines a $\mcL_1$ reconstruction term and a perceptual term \(\mcL_{\text{SSIM}}\) that is based on the Structural Similarity Index Measure (SSIM)~\cite{Wang2003}
\begin{equation}
\label{eq:3dgs_loss}
  \mcL = (1-\lambda) \mcL_1 + \lambda\,\mcL_{\text{SSIM}}.
\end{equation}

To fully utilize real-world event camera data, undistortion operation is performed as the accumulated image $D$ is computed. We account for this by only incurring the loss where events were actually accumulated. This allows supervising the 3D representation more efficient. 

%% file: sec/5_experiments.tex
\begin{figure*}[t]
  \centering
  \includegraphics[width=13.5cm]{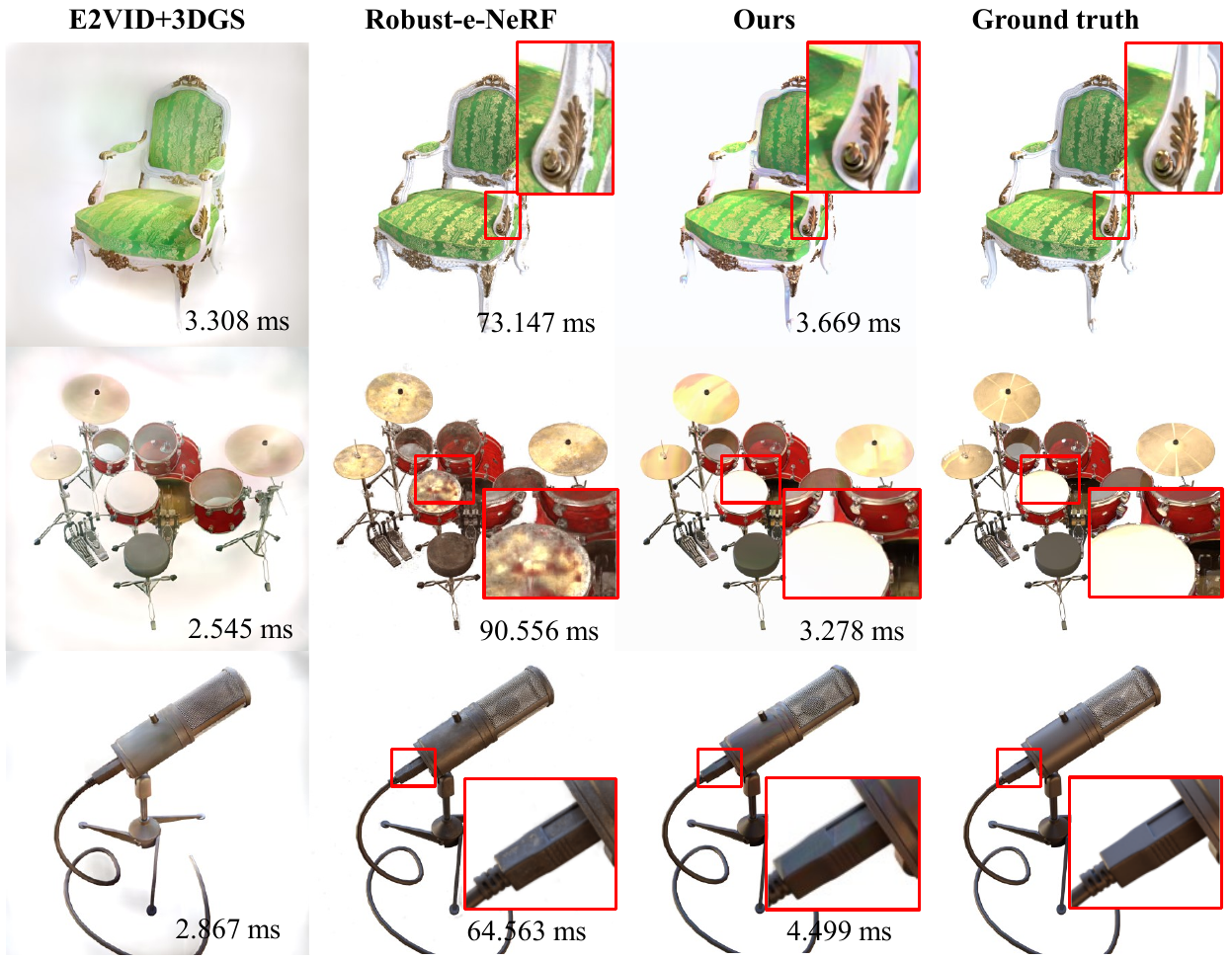}
  \caption{Generated Synthetic images comparing our work, event-based NeRF, and E2VID+3DGS qualitatively, with rendering times shown at the bottom of each image. Our work and Robust-e-NeRF outperform others in most scenes. In some cases, e.g., the drums scene, our work performs better and achieves significantly faster rendering times than conventional Robust-e-NeRF composed of InstantNGP.}
  \label{fig:qualitatives}
\end{figure*}

\begin{table*}[tb]
\setlength{\tabcolsep}{3.6pt}
  \centering
  \caption{Generated synthetic scenes in comparison between our work, previous event-based NeRF and E2VID+3DGS quantitatively. We supervised our model with randomly initialized points.}
  \begin{tabular}{llcccccccc}
    \toprule
    \multicolumn{1}{c}{} & \multicolumn{1}{c}{} & \multicolumn{7}{c}{Synthetic Scene} &  \\ \cmidrule(lr){3-9}
    \multicolumn{1}{c}{\multirow{-2}{*}{Metric}} & \multicolumn{1}{c}{\multirow{-2}{*}{Method}} & \texttt{chair} & \texttt{drums} & \texttt{ficus} & \texttt{hotdog} & \texttt{lego} & \texttt{materials} & \texttt{mic} & \multirow{-2}{*}{Mean} \\
                                         \midrule
                                         & E2VID $+$ 3DGS         & 21.39 & 19.86 & 19.90 & 15.55 & 18.17 & 20.08 & 20.10 & 19.29 \\
                                         & Robust \textit{e}-NeRF & {\bf30.24} & \underline{23.15} & {\bf30.71} & {\bf28.07} & \underline{27.34} & \underline{24.98} & \underline{32.87} & {\bf28.19} \\
                                         \multirow{-3}{*}{PSNR $\uparrow$}
                                         & Ours                   & \underline{28.69} & {\bf25.81} & \underline{29.90} & \underline{22.91} & {\bf29.22} & {\bf27.16} & {\bf33.27} & \underline{28.14} \\
                                         \midrule
                                         & E2VID $+$ 3DGS         & 0.934 & 0.915 & 0.922 & 0.897 & 0.895 & 0.901 & 0.957 & 0.917 \\
                                         & Robust \textit{e}-NeRF & {\bf0.958} & \underline{0.897} & {\bf0.971} & {\bf0.953} & \underline{0.934} & \underline{0.923} & \underline{0.981} & \underline{0.945} \\
                                         \multirow{-3}{*}{SSIM $\uparrow$}
                                         & Ours                   & \underline{0.953} & {\bf0.947} & \underline{0.966} & \underline{0.940} & {\bf0.945} & {\bf0.936} & {\bf0.986} & {\bf0.953} \\
                                         \midrule
                                         & E2VID $+$ 3DGS         & 0.076 & 0.094 & 0.108 & 0.208 & 0.145 & 0.125 & 0.069 & 0.118 \\
                                         & Robust \textit{e}-NeRF & {\bf0.040} & \underline{0.091} & {\bf0.022} & {\bf0.095} & \underline{0.074} & {\bf0.052} & \underline{0.029} & \underline{0.057} \\
                                         \multirow{-3}{*}{LPIPS $\downarrow$}
                                         & Ours                   & \underline{0.047} & {\bf0.052} & \underline{0.028} & \underline{0.098} & {\bf0.055} & \underline{0.060} & {\bf0.015} & {\bf0.051} \\
                                         \bottomrule
  \end{tabular}
\label{tab:quantitative_synthetic}
\end{table*}

\begin{figure*}[tb]
  \centering
  \includegraphics[width=0.7\linewidth]{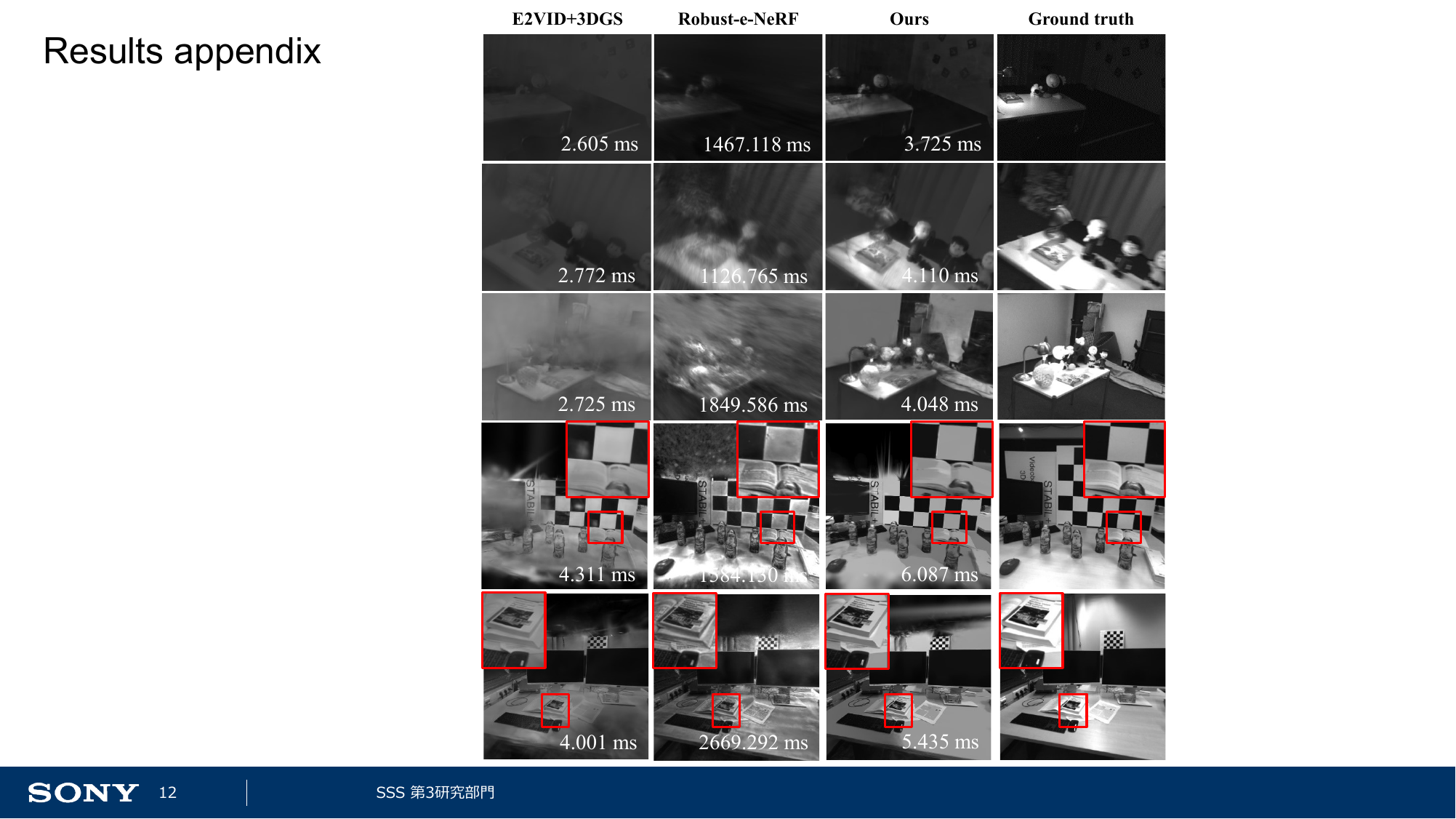}
  \caption{Qualitative comparisons of the images generated by our method, event-based NeRF and E2VID+3DGS show that our technique appears to recover details in 5 real scenes. First three rows(\text{07\_ziggy\_and\_fuzz\_hdr}, \text{08\_peanuts\_running} and \text{11\_all\_characters}) are EDS dataset and last two rows are TUM-VIE dataset(\text{mocap-1d-trans} and \text{mocap-desk2}).}
  \label{fig:qualitatives_real}
\end{figure*}

\begin{table*}[tb]
\setlength{\tabcolsep}{9.0pt}
  \centering
  \caption{Real scenes in comparison between our work, previous event-based NeRF and E2VID+3DGS quantitatively.}
  \begin{tabular}{llcccccc}
    \toprule
    \multicolumn{1}{c}{} & \multicolumn{1}{c}{} & \multicolumn{5}{c}{Real Scene} &  \\ \cmidrule(lr){3-7}
    \multicolumn{1}{c}{\multirow{-2}{*}{Metric}} & \multicolumn{1}{c}{\multirow{-2}{*}{Method}} & \texttt{03} & \texttt{07} & \texttt{08} & \texttt{11} & \texttt{13} & \multirow{-2}{*}{Mean} \\
                                         \midrule
                                         & E2VID $+$ 3DGS         & 15.67 & \underline{15.05} & 14.03 & 13.83 & \underline{18.96} & 15.51 \\
                                         & Robust \textit{e}-NeRF & \underline{19.19} & 14.78 & \underline{14.75} & \underline{14.43} & 18.10 & \underline{16.25} \\
                                         \multirow{-3}{*}{PSNR $\uparrow$}
                                         & Ours                   & {\bf20.78} & {\bf19.14} & {\bf17.53} & {\bf17.79} & {\bf19.05} & {\bf18.86} \\
                                         \midrule
                                         & E2VID $+$ 3DGS         & 0.716 & 0.689 & 0.642 & \underline{0.691} & 0.723 & 0.692 \\
                                         & Robust \textit{e}-NeRF & {\bf0.846} & \underline{0.815} & \underline{0.735} & 0.569 & \underline{0.729} & \underline{0.739} \\
                                         \multirow{-3}{*}{SSIM $\uparrow$}
                                         & Ours                   & \underline{0.835} & {\bf0.816} & {\bf0.745} & {\bf0.789} & {\bf0.774} & {\bf0.792} \\
                                         \midrule
                                         & E2VID $+$ 3DGS         & \underline{0.266} & \underline{0.378} & {\bf0.402} & \underline{0.415} & \underline{0.415} & \underline{0.375} \\
                                         & Robust \textit{e}-NeRF & 0.324 & 0.476 & 0.567 & 0.700 & 0.650 & 0.543 \\
                                         \multirow{-3}{*}{LPIPS $\downarrow$}
                                         & Ours                   & {\bf0.239} & {\bf0.351} & \underline{0.424} & {\bf0.391} & {\bf0.407} & {\bf0.363} \\
                                         \bottomrule
  \end{tabular}
\label{tab:quantitative_real}
\end{table*}

\section{Experiments}
\label{sec:experiments}

\subsection{Datasets}
\label{sec:scenes}
Our evaluation of 3D Gaussian Splatting from a moving event camera encompasses both synthetic and real-world scenes. Our method assumes that we can access the intrinsic camera matrix, its lens distortion parameters, and the camera's constant-rate poses, sampled at a high frequency to ensure precise interpolation for the event camera at any chosen time point.
Estimation of camera poses at any given timestep is crucial for precise scene reconstruction from event data, as events can asynchronously occur at arbitrary timesteps between the constant-rate sampled camera poses.

\subsubsection{Synthetic Scenes}
The synthetic dataset used in our experiments consists of event data streams obtained from ESIM~\cite{pmlr-v87-rebecq18a}, initially introduced in~\cite{low2023robust}, encompassing seven synthetic scenes with a white background. These simulated event data streams feature a diverse range of photorealistic objects with complex structures. The event camera in these synthetic scenes moves around the objects, providing diverse geometric and lighting conditions. This setup allows for a comprehensive evaluation of the robustness of our method against the baselines in a controlled environment.

\subsubsection{Real-World Scenes}

The experiments on real scenes were conducted using subsets of the EDS~\cite{Hidalgo2022cvpr}, and TUM-VIE~\cite{9636728} datasets. From EDS, we utilize the sequences \text{03\_rocket\_earth\_dark}, \text{07\_ziggy\_and\_fuzz\_hdr}, \text{08\_peanuts\_running}, \text{11\_all\_characters} and \text{13\_airplane}. From TUM-VIE, we utilize the sequences \text{mocap-1d-trans} and \text{mocap-desk2}.
These sequences primarily capture various objects in 360-degree scenes and on a desk from a forward-facing perspective. These real sequences are relatively well-suited for novel view synthesis and were recorded using the high-resolution event sensor, the Prophesee Gen 3 in the EDS and the Prophesee Gen 4 in the TUM-VIE dataset, respectively.

\subsection{Baselines}
We benchmarked our method against Robust-e-NeRF and E2VID~\cite{rebecq2019high} + 3D Gaussian Splatting (E2VID+3DGS). Robust-e-NeRF incorporates InstantNGP, which is significantly faster compared to traditional NeRF methods, but still falls short compared to 3DGS. E2VID generates RGB images from an event data stream, followed by feeding the images into original 3D Gaussian Splatting.

\subsection{Metrics}
\label{sec:metrics}
We utilized commonly used metrics to evaluate novel view synthesis, including Peak Signal-to-Noise Ratio (PSNR), Structural Similarity Index Measure (SSIM), and Learned Perceptual Image Patch Similarity (LPIPS~\cite{zhang2018unreasonable}) with VGG16, to comprehensively assess the quality of rendered images. It's worth noting that SSIM and LPIPS are perceptual metrics introduced to align more closely with human preference, with LPIPS specifically tailored to mimic human perception. These metrics address issues associated with PSNR, such as favoring blurry images. Additionally, we measured and compared the rendering speeds of our method, Robust-e-NeRF, and E2VID+3DGS.

Since event cameras capture variations in log-radiance rather than absolute log-radiance values, the predicted intensity~$I(t)$ from the 3D Gaussian Splatting has an unknown offset. To rectify this limitation, a linear color transformation is designed to adjust our predictions in the logarithmic domain~\cite{rudnev2023eventnerf, low2023robust}.
This transformation is both necessary and adequately effective for aligning our predictions with the reference data. It ensures that the reconstructed intensity values are properly calibrated and aligned with the observed event data.

\subsection{Results and Evaluation}
\label{results}
After completing the training of our model on each scene using event sequence data, the model acquires a set of 3D Gaussian ellipsoids representing the scene. By rasterizing these Gaussians, we are able to render various arbitrary views of each scene in real-time. The renderings of novel views, obtained from held-out views or test views provided by the datasets, are then used for evaluations.

\subsubsection{Synthetic Scenes}
\label{synthetic_results}
We evaluate our methodology on a synthetic dataset using the metrics described in~\cref{sec:metrics}. Quantitative results are shown in~\cref{tab:quantitative_synthetic}, and visual samples and rendering times in~\cref{fig:qualitatives}. Rendering times are provided in milliseconds (ms). 
Our method achieves higher average SSIM and compared to previous works. A high SSIM value indicates structural similarity between generated and reference images, underscoring our model's accuracy in capturing essential scene geometry. Additionally, a low LPIPS score reflects perceptual quality, showing the produced images are both realistic and visually engaging. The elevated scores demonstrate that our novel view synthesis model faithfully reproduces the visual and perceptual attributes of the original scene.
Moreover, the rendering speed of both our method and E2VID+3DGS significantly surpass that of Robust-e-NeRF. Therefore, our work maintains superior quality and speed compared to Robust-e-NeRF, highlighting its effectiveness in both fidelity and performance.

As shown in~\cref{fig:qualitatives}, the quality of rendered images remains high in scenes featuring drums and microphones, achieving novel view synthesis without sacrificing quality in other scenes as well.

\subsubsection{Real Scenes}
\label{real_results}

The EDS dataset is the only one available for quantifying the performance of novel view synthesis. Therefore, we primarily conducted quantitative and qualitative assessments, as shown~\cref{tab:quantitative_real} and~\cref{fig:qualitatives_real}.
Our method successfully recovers fine details in all sequences and contrasts in the \text{mocap-1d-trans} image, particularly. This is particularly noticeable in the images containing the checkerboard and the cover of a book.
Additionally, it effectively smooths uniform areas such as the checkerboard and the desk surface in the \text{mocap-1d-trans} and \text{mocap-desk2}. This smoothness likely results from the accumulation of event streams, which may have enhanced the signal-to-noise ratio due to the aggregation process. The visible artifacts near the borders of all synthesized views are attributed to the relatively narrow field of view of the event cameras.

\subsection{Ablations}
\begin{figure}[t]
  \centering
  \includegraphics[width=\linewidth]{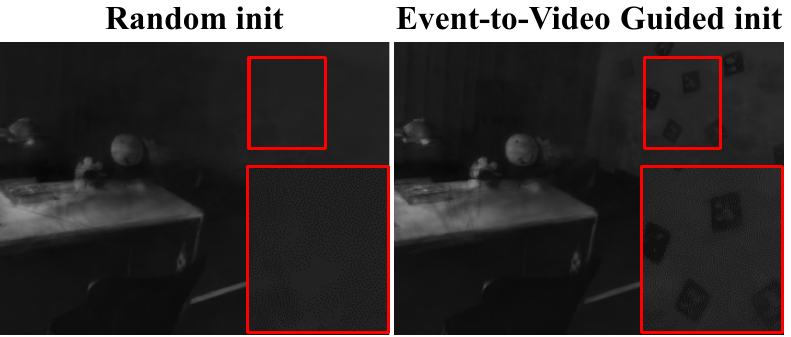}
  \caption{Qualitative comparisons of the images generated by random initialization and event-to-video guided initialization are shown in~\text{07\_ziggy\_and\_fuzz\_hdr} scene in EDS dataset.}
  \label{fig:randominit_e2vidinit}
\end{figure}

To demonstrate the effectiveness of event-to-video guided initialization and camera trajectory estimation with cubic spline interpolation, we conducted ablation experiments by separately adding 1) event-to-video guided initialization and 2) both the initialization and cubic spline interpolation. As shown in~\cref{tab:ablation}, event-to-video guided initialization outperforms random initialization. Furthermore,~\cref{fig:randominit_e2vidinit} shows that our initialization approach results in better image quality (e.g. backgrounds) compared to random initialization. Adding cubic spline interpolation further enhances performance.

\begin{table}[tb]
  \centering
  \caption{Ablation results for the event-to-video guided initialization and cubic spline interpolation.}
  \begin{tabular}{c|ccc}
    \toprule
    \textbf{Method} & \textbf{PSNR$\uparrow$} & \textbf{SSIM$\uparrow$} & \textbf{LPIPS$\downarrow$} \\ \hline
    random init        & 18.02 & 0.767 & 0.397 \\ 
    guided init        & \underline{18.75} & \underline{0.788} & {\bf0.359} \\ 
    guided init + cubic & {\bf18.86} & {\bf0.792} & \underline{0.363} \\ 
    \bottomrule
  \end{tabular}
\label{tab:ablation}
\end{table}

\subsection{Limitations}
\label{limitations}
Our method, though effective, has some limitations. It is designed exclusively for novel view synthesis in static scenes and cannot handle scenarios with dynamic moving objects, which we consider to be an exciting avenue for future research. Additionally, our approach trains 3D Gaussian Splatting on event-accumulated images between two viewpoints, representing relative intensity images. Consequently, similar to other event-based view synthesis methods, the model cannot directly estimate absolute intensity images and requires a linear transformation with evaluation data as a reference. This linear transformation is required only during inference and does not impact the training.

%% file: sec/6_conclusion_and_futurework.tex
\section{Conclusion and Future Work}
In this work, we introduce a novel method for 3D Gaussian Splatting from moving event cameras. Our approach achieves the higher visual fidelity and better performance of novel view synthesis in static scenes.
We distill geometric priors from the event-to-video model for initialization and leverage effective camera trajectory interpolation. Thus, EventSplat yields improved metrics compared to existing event-based NeRF approaches.
Moreover, the incorporation of 3D Gaussian Splatting enables our model to render scenes at significantly higher frame rates. This enhances the practicality of real-time novel view synthesis with event cameras. Our work is thus particularly useful in scenarios where conventional imaging techniques are inadequate.

%% file: sec/appendix.tex
\clearpage
\newpage
\appendix

\section{Implementation Details}
\label{app:alg_details}
\subsection{Algorithm}
Our optimization and densification algorithm is shown in~\cref{alg:optimization}. All modifications compared to the original Gaussian Splatting process~\cite{kerbl20233d} are highlighted in {\color{uoftgreen} green}.

\begin{algorithm}[!h]
\caption{Optimization and Densification\\
$w$, $h$: width and height of the training images}
\label{alg:optimization}
\begin{algorithmic}
  \State {\color{uoftgreen} $M \gets$  Event-to-VideoGuidedPoints()}	\Comment{Positions}
  \State $S, C, A \gets$ InitAttributes()  \Comment{Covariances, Colors, Opacities}
  \State $i \gets 0$	\Comment{Iteration Count}
  \While{not converged}
    \State {\color{uoftgreen}
      $k_{start}, k_{end}, D \gets$ GenerateTrainingView() }
    \State {\color{uoftgreen}
      $I_1 \gets$ 
      Rasterize($M$, $S$, $C$, $A$, $k_{start}$)}
    \State {\color{uoftgreen}
      $I_2 \gets$ 
      Rasterize($M$, $S$, $C$, $A$, $k_{end}$)}
    \State {\color{uoftgreen}$\hat{L}_1 \gets$ Log(Remosaicing($I_1$))}
    \State {\color{uoftgreen}$\hat{L}_2 \gets$ Log(Remosaicing($I_2$))}
    \State {\color{uoftgreen}
      $\mcL \gets$ Loss($\hat{L}_2-\hat{L}_1$, $D$)}
      \Comment{Loss}
    \State $M$, $S$, $C$, $A$ $\gets$ Adam($\nabla \mcL$) \Comment{Backprop \& Step}

			\If{IsRefinementIteration($i$)}
			\ForAll{Gaussians $(\mu, \Sigma, c, \alpha)$ $\textbf{in}$ $(M, S, C, A)$}
			\If{$\alpha < \epsilon$ or IsTooLarge($\mu, \Sigma)$}	\Comment{Pruning}
			\State RemoveGaussian()	
			\EndIf
			\If{$\nabla_p L > \tau_p$} \Comment{Densification}
			\If{$\|S\| > \tau_S$}	\Comment{Over-reconstruction}
			\State SplitGaussian($\mu, \Sigma, c, \alpha$)
			\Else								\Comment{Under-reconstruction}
			\State CloneGaussian($\mu, \Sigma, c, \alpha$)
			\EndIf	
			\EndIf
			\EndFor		
			\EndIf
			\State $i \gets i+1$
			\EndWhile
		\end{algorithmic}
	\end{algorithm}

\subsection{Hyper-parameters and Optimizations}
\label{app:params_opts}
Our approach adopts original 3D Gaussian Splatting as the backbone as it allows for high quality view synthesis with high-speed rendering.
The Gaussian Model is initialized with spherical harmonics degree and several parameters, including xyz coordinates, features, scaling, rotation, and opacity. The model sets up essential functions for covariance, opacity, and rotation activations.
The model includes functions to densify and prune Gaussians based on gradient thresholds and opacity. This ensures efficient use of computational resources by adding new Gaussians where needed and removing those that are not contributing effectively.
Training utilizes the similar optimization strategies and hyper-parameter settings originally proposed for 3D Gaussian Splatting including position, feature, opacity, scaling, and rotation. The learning rate is scheduled to adjust dynamically during training. The only opacity learning rate was changed from 0.05 to 0.01 to make the training more stable. The instability seems to result from the 3D Gaussian Splatting model being supervised from multi-view points with different accumulation lengths.

\subsubsection{Contrast threshold}
Both Robust-e-NeRF and our method were co-optimized and trained with the symmetric contrast thresholds initialized at~$\nicefrac{C_{+1}}{C_{-1}}=1.0$ (more precisely set at~$C_{-1}=0.25$) in synthetic datasets and the EDS dataset, and asymmetric contrast threshold initialized at~$\nicefrac{C_{+1}}{C_{-1}}=1.458$ (set at~$C_{-1}=0.25$)\cite{low2023robust} in the TUM-VIE dataset.

\subsection{Experiment Setup}
Our research and development efforts are deeply rooted in the principles of 3D Gaussian Splatting~\cite{kerbl20233d} methodology. In pursuit of advancing these technologies, we trained our models for more than 30k iterations (set at 40k iterations). This training was conducted on a NVIDIA GeForce RTX4090 GPU. Training time of synthetic scenes take 1-2 hours and that of real scenes take 1-3 hours at 40k iterations.

\section{Additional Experimental Results}
\label{app:other_experiments}

\subsection{Qualitative Results in Synthetic Scenes}
\begin{figure*}[tb]
  \centering
  \includegraphics[height=21.0cm]{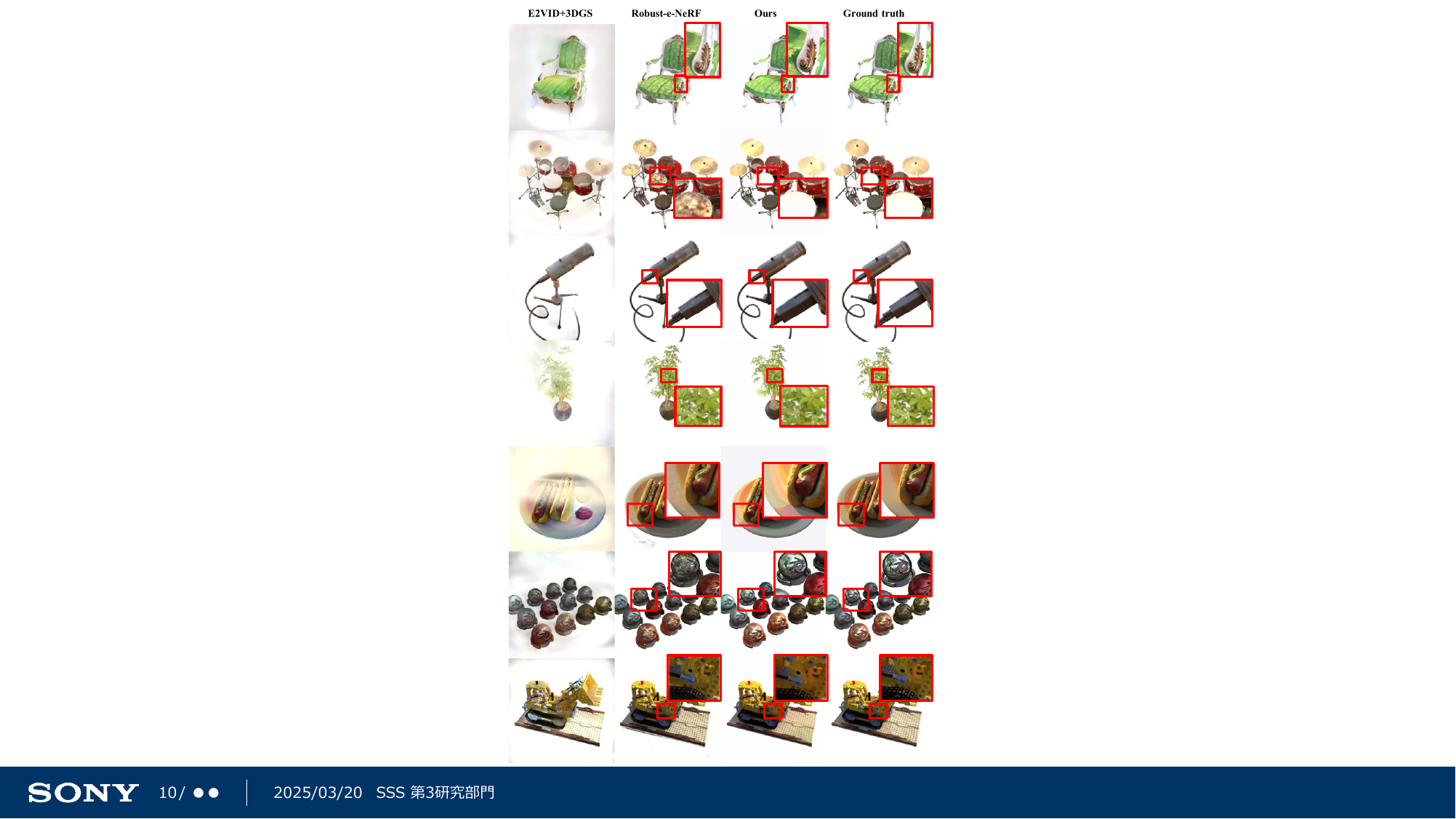}
  \caption{Generated images are shown, qualitatively comparing our work, event-based NeRF, and E2VID+3DGS in all synthetic scenes.}
  \label{fig:qualitatives_appendix}
\end{figure*}

~\cref{fig:qualitatives_appendix} shows the quantitative results of all methods for all seven synthetic scenes. The qualitative results are similar to the quantitative evaluation numbers as shown in~\cref{tab:quantitative_synthetic}. In the drums, lego, materials, and mic scenes, fine details seem to be well reconstructed. The chair and ficus reconstruction results appear to be similar details. In the hotdog case, it seems that the images produced by our method are not as well reconstructed compared to Robust-e-NeRF.

\subsection{Qualitative Results on Different Synthetic Datasets}
\begin{figure}[tb]
  \centering
  \includegraphics[width=\linewidth]{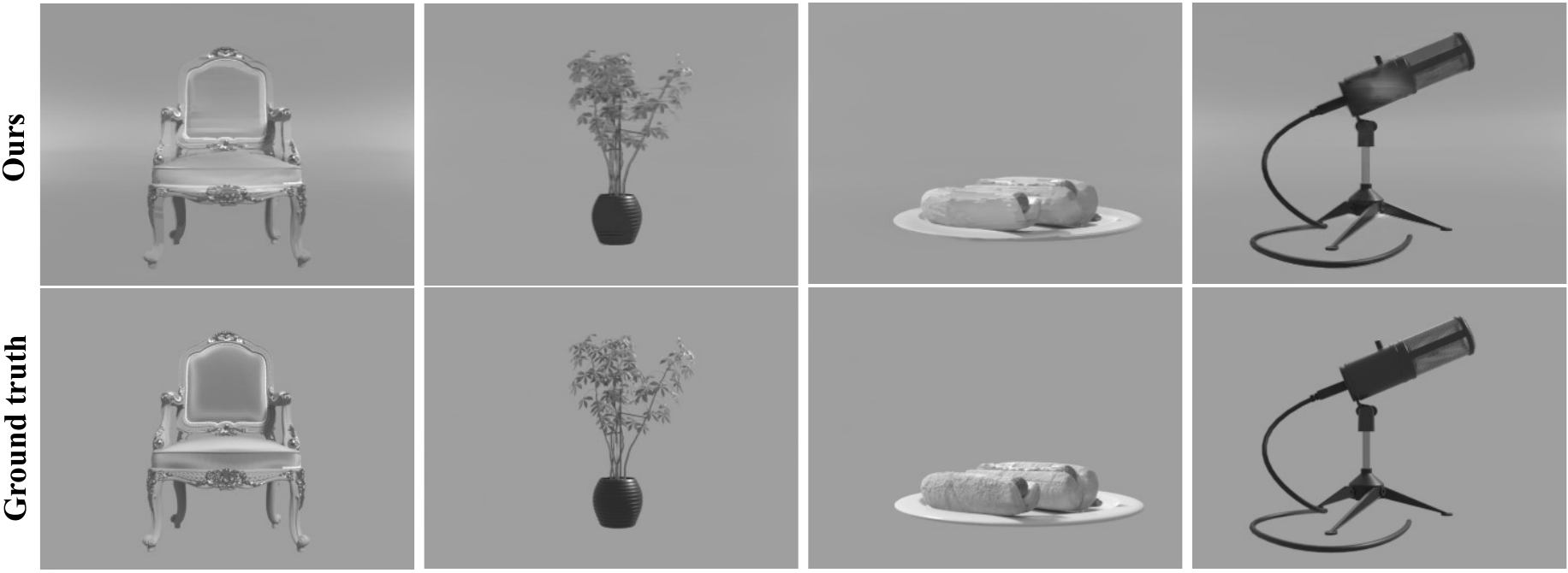}
  \caption{Generated images qualitatively comparing our method with ground truth across 4 synthetic scenes.}
  \label{fig:qualitatives_synthetic_eventnerf_appendix}
\end{figure}

We evaluated our method on the same scenes used in EV-GS~\cite{wu2024ev} from EventNeRF dataset~\cite{rudnev2023eventnerf} to compare grayscale results.
We computed the mean values across 4 scenes(chair, ficus, hotdog and mic).
As shown in~\cref{tab:quantitative_synthetic_eventnerf_appendix}, our approach outperforms EV-GS in terms of PSNR and SSIM.
Furthermore,~\cref{fig:qualitatives_synthetic_eventnerf_appendix} shows qualitative results for 4 synthetic scenes in grayscale, demonstrating that the generated images are reconstructed effectively.

\begin{table}[tb]
  \centering
  \caption{Quantitative evaluation of mean values across the 4 synthetic scenes from~\cite{wu2024ev}.}
  \begin{tabular}{lcc|cc}
    \toprule
    {Metric} & \multicolumn{2}{c|}{PSNR $\uparrow$} & \multicolumn{2}{c}{SSIM $\uparrow$} \\
    & Ours & EV-GS & Ours & EV-GS \\
    \midrule
    Mean & \textbf{29.48} & 26.6 & \textbf{0.959} & 0.925 \\
    \bottomrule
  \end{tabular}
  \label{tab:quantitative_synthetic_eventnerf_appendix}
\end{table}

\subsection{Qualitative Results in Real-World Scenes}
\begin{figure*}[tb]
  \centering
  \includegraphics[height=20.0cm]{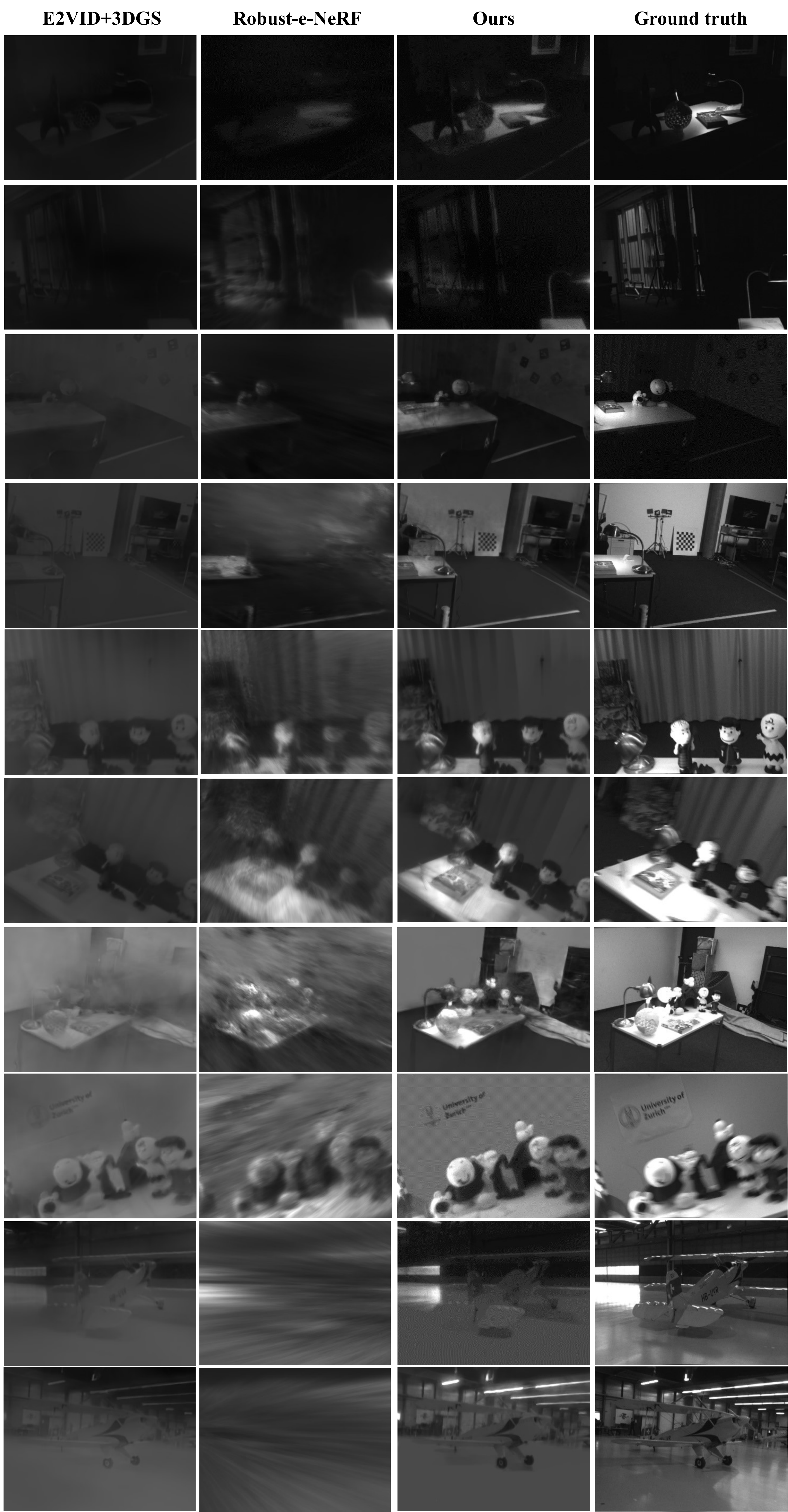}
  \caption{For each scene in the EDS dataset, we show generated images from two viewpoints alongside the ground truth image, comparing our work with event-based NeRF and E2VID+3DGS.}
  \label{fig:qualitatives_real1_appendix}
\end{figure*}

\begin{figure*}[tb]
  \centering
  \includegraphics[width=14cm]{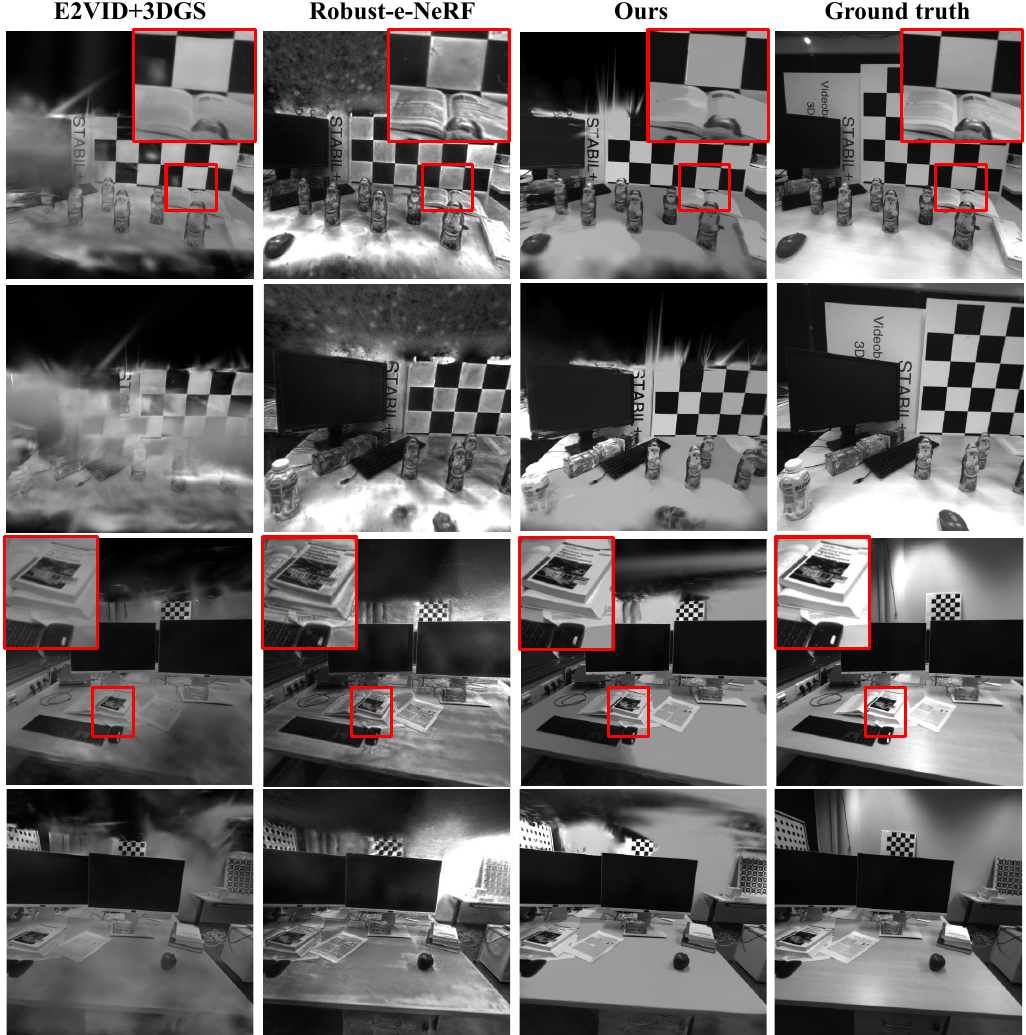}
  \caption{For each scene in TUM-VIE dataset, we show generated images from two viewpoints alongside the ground truth image, comparing our work with event-based NeRF and E2VID+3DGS.}
  \label{fig:qualitatives_real2_appendix}
\end{figure*}

~\cref{fig:qualitatives_real1_appendix} and ~\cref{fig:qualitatives_real2_appendix} present additional quantitative results for the scenes \text{03\_rocket\_earth\_dark}, \text{07\_ziggy\_and\_fuzz\_hdr}, \text{08\_peanuts\_running}, \text{11\_all\_characters} and \text{13\_airplane}, as well as for the \text{mocap-1d-trans}, \text{mocap-desk2} scenes, respectively. Our method demonstrates the ability to reconstruct fine detail in both real-world data.
However, there is still room for improvement in the quality of reconstruction for some real-world scenes, particulary concerning floating point clouds and the back wall.

